\documentclass[final]{opt2025}

\usepackage[utf8]{inputenc} 
\usepackage[T1]{fontenc}    
\usepackage{enumitem}
\usepackage{hyperref}       
\usepackage{url}            
\usepackage{booktabs}       
\usepackage{amsfonts}       
\usepackage{microtype}      
\usepackage{xcolor}         

\usepackage{amsmath, amssymb, amsfonts}
\usepackage{algorithm}
\usepackage[noend]{algpseudocode}
\usepackage{adjustbox}
\usepackage{wrapfig}
\usepackage{graphicx}

\newcommand{\E}{\mathbb{E}}
\newcommand{\R}{\mathbb{R}}

\newcommand{\dataset}{\mathcal{D}}
\newcommand{\labels}{\mathcal{T}} 
\newcommand{\privileged}{\mathcal{P}} 
\newcommand{\nonprivileged}{\bar{\mathcal{P}}} 
\newcommand{\confusing}{S_{il}} 
\newcommand{\model}{m} 
\newcommand{\params}{\mathbf{w}} 
\newcommand{\Params}{\{\mathbf{w}_t \vert t \in \mathcal{T}\}} 
\newcommand{\refparams}{\hat{\mathbf{w}}} 
\newcommand{\Refparams}{\{\hat{\mathbf{w}}_t \vert t \in \mathcal{T}\}} 
\newcommand{\loss}{\mathcal{L}}
\newcommand{\baseloss}{\ell} 
\newcommand{\policy}{\pi}
\newcommand{\refpolicy}{\pi_{\text{ref}}}
\newcommand{\attributes}{\mathcal{A}}

\title[FairPO: Fair Preference Optimization for Multi-Label Learning]{FairPO: Fair Preference Optimization for Multi-Label Learning}

\optauthor{%
\Name{Soumen Kumar Mondal} \Email{soumenkm@iitb.ac.in}\\
\Name{Prateek Chanda$^{\dagger}$} \Email{prateek.chanda@iitb.ac.in}\\
\Name{Akshit Varmora$^{\dagger}$} \Email{akshit.varmora@iitb.ac.in}\\
\Name{Ganesh Ramakrishnan} \Email{ganramk@iitb.ac.in}\\
\addr IIT Bombay, India}

\begin{document}

\maketitle
\footnotetext{$^{\dagger}$ Equal contribution}

\begin{abstract}
Multi-label classification (MLC) often suffers from performance disparities across labels. We propose \textbf{FairPO}, a framework combining preference-based loss and group-robust optimization to improve fairness by targeting underperforming labels. FairPO partitions labels into a \textit{privileged} set for targeted improvement and a \textit{non-privileged} set to maintain baseline performance. For privileged labels, a DPO-inspired preference loss addresses hard examples by correcting ranking errors between true labels and their confusing counterparts. A constrained objective maintains performance for non-privileged labels, while a Group Robust Preference Optimization (GRPO) formulation adaptively balances both objectives to mitigate bias. We also demonstrate FairPO's versatility with reference-free variants using Contrastive (CPO) and Simple (SimPO) Preference Optimization\footnote{Our code is available at GitHub: \url{https://github.com/soumenkm/FairPO}}.
\end{abstract}


\section{Introduction}
\paragraph{The Challenge of Performance Disparity in MLC.}Standard multi-label classification (MLC) models assume all labels are equal, minimizing an aggregate loss over the entire label set \citep{kiritchenko2006hierarchical, DBLP:journals/corr/WeiXLZ0G16, schietgat2010predicting}. This assumption fails in real-world scenarios where data asymmetries cause significant performance disparities. These disparities stem from several factors, including \textbf{class imbalance}, where models favor frequent labels over rare ones \citep{Charte2015AddressingII, 6471714}, and varying \textbf{real-world importance}, such as distinguishing a critical disease from a benign one \citep{rajpurkar2017chexnetradiologistlevelpneumoniadetection, 10368872}. Disparities also arise from \textbf{semantic complexity}, where models learn simple concepts but fail on nuanced ones. Consequently, training becomes dominated by the easy majority, yielding systems that are unreliable for the very cases that matter most \citep{swayamdipta2020datasetcartographymappingdiagnosing, pleiss2020identifyingmislabeleddatausing}.

\paragraph{Limitations of Conventional Methods.}
Conventional methods are limited by the aggregate nature of the Binary Cross-Entropy (BCE) loss, which allows the learning signal for underrepresented labels to be drowned out by easy ones \citep{bce_loss, Charte2015AddressingII}. While solutions like Focal Loss help, they are fundamentally restricted because they treat each label's prediction in isolation \citep{lin2018focallossdenseobject}. They do not teach the model to perform \textit{discrimination} by resolving confusion between a true label and a specific negative alternative. This reveals a gap for a framework that can learn \textit{relative preferences}, improving performance on difficult labels while maintaining it for others. A preference-based objective forces the model to learn that a true label's score is explicitly \textit{preferred} over that of a confusing competitor.

\paragraph{Our Approach: The FairPO Framework.}
We introduce \textbf{FairPO} (Fair Preference Optimization), a framework to manage performance disparities by partitioning labels into a \emph{privileged} group for targeted improvement and a \emph{non-privileged} group for performance maintenance. We prioritize learning on the \emph{privileged} group using a preference-based objective, a novel adaptation of techniques from generative model alignment for a discriminative task. For each privileged true label, FairPO dynamically identifies its \textit{confusing counterparts}, incorrect labels with misleadingly high scores and trains the model to prefer the true label, creating a large discriminative margin. Concurrently, a constrained objective safeguards the \emph{non-privileged} group against performance degradation relative to a reference model. These competing objectives are dynamically balanced using a Group Robust Preference Optimization (GRPO) formulation \citep{ramesh2024group}.

\begin{wrapfigure}{r}{0.3\textwidth} 
    \vspace{-20pt} 
    \centering
    \includegraphics[width=\linewidth]{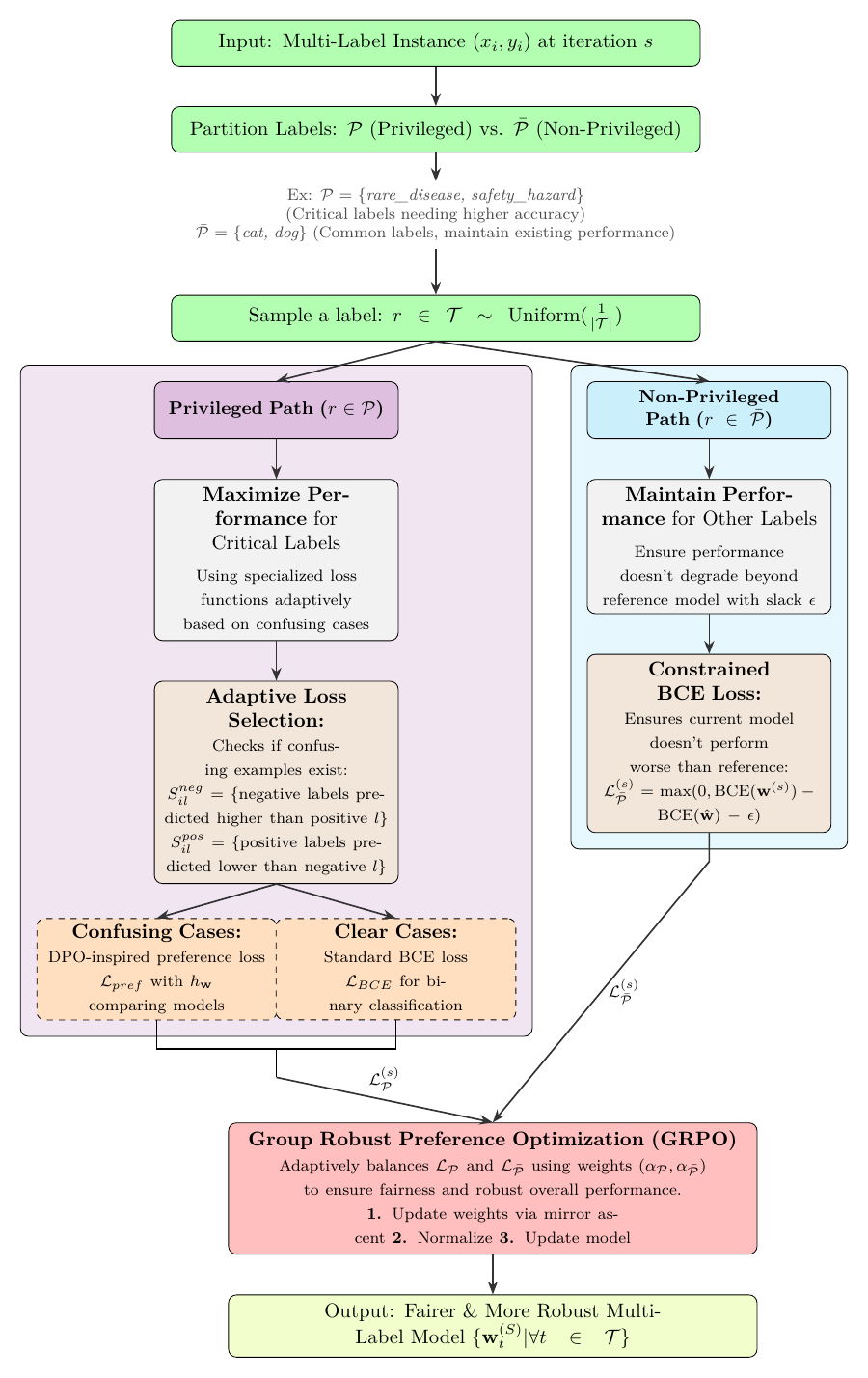}
    \caption{FairPO Framework}
    \label{fig:fairpo_overview}
    \vspace{-10pt} 
\end{wrapfigure}

\paragraph{Application and Contributions.}
Our general FairPO framework can address any MLC task requiring label prioritization due to factors like class imbalance or real-world importance. To demonstrate its effectiveness, we apply it to \textbf{class imbalance}, defining infrequent tail labels as the \emph{privileged} group and frequent head labels as the \emph{non-privileged} one. Our main contributions are:
\begin{enumerate}
\item We propose \textbf{FairPO}, a novel framework for targeted performance improvement in MLC applicable to various sources of disparity.
\item We are the first to adapt preference optimization techniques \citep{meng2024simposimplepreferenceoptimization, xu2024contrastivepreferenceoptimizationpushing} from generative modeling to create discriminative objectives for MLC.
\item We successfully apply GRPO to solve the trade-off between objectives for the \emph{privileged} and \emph{non-privileged} groups, a novel application in this context.
\end{enumerate}
Experiments show FairPO significantly boosts performance on infrequent labels by up to 3.59\% mAP while robustly maintaining performance on frequent ones, outperforming several baselines.

\section{Methodology: The FairPO Framework}
\label{sec:methodology}

Our methodology builds upon a foundation of preference optimization techniques. To provide the necessary background on these methods, namely DPO, CPO, SimPO, and GRPO, we have moved the detailed review of their core formulations to Appendix \ref{sec:preliminaries}.

\subsection{Problem Setup and Fairness Goals}

Let us consider a standard multi-label classification setting with a dataset $\dataset = \{ (\mathbf{x}_i, y_i) \}_{i=1}^N$, where $\mathbf{x}_i$ is an input instance and $y_i \in \{0, 1\}^{|\labels|}$ is the corresponding multi-label vector over a universe of $|\labels|$ labels. The goal is to train a model, parameterized by weights $\params$, that produces a set of per-label scores $\model(\mathbf{x}_i; \params_t)$ for each label $t \in \labels$. Our framework also utilizes a pre-trained reference model with parameters $\refparams_t$.

Conventional methods often train such models by minimizing a single, aggregate loss where each label contributes equally. However, as motivated in our introduction, real-world applications frequently demand a more nuanced approach where certain labels are prioritized. Our framework, FairPO, is explicitly designed for these scenarios. The core idea is to partition the total label set $\labels$ into two disjoint subsets:
\begin{itemize}
    \item A \textbf{\emph{privileged} group} $\privileged \subset \labels$, which contains labels for which we seek to significantly enhance the model's performance.
    \item A \textbf{\emph{non-privileged} group} $\nonprivileged = \labels \setminus \privileged$, which contains the remaining labels, for which our objective is to robustly maintain at least a baseline level of performance.
\end{itemize}

This partitioning is a key strength of our framework's general design. The criteria for assigning labels to the \emph{privileged} group are flexible and can be adapted to the specific problem at hand, such as using label frequency for class imbalance, domain knowledge for real-world importance, or annotation consistency for data quality issues. In this work, to demonstrate FairPO's efficacy, we focus on the class imbalance problem, where the \emph{privileged} group consists of the least frequent labels. We defer the precise details of this setup to Section~\ref{sec:experiments}.

\subsection{FairPO Objectives}
\label{subsec:fairpo_objectives}

FairPO's methodology is built on three core components: a conditional objective for privileged labels, a constrained objective for non-privileged labels, and a robust optimization framework to balance them. The detailed mathematical formulations for each component are provided in Appendix~\ref{app:fairpo_detailed_method}.

\textbf{Objective for Privileged Labels ($l \in \privileged$):} To improve performance on critical labels, we employ a conditional objective designed to target hard-to-discriminate cases. When a \textit{confusing set} of labels exists for an instance (i.e., high-scoring negatives or low-scoring positives), a preference loss inspired by DPO \citep{rafailov2023direct} is applied to enforce correct relative rankings. In the absence of such confusing examples, the objective reverts to a standard BCE loss, which acts as a crucial fallback for stable training. Our framework is versatile and also supports reference-free preference loss variants inspired by CPO \citep{xu2024contrastivepreferenceoptimizationpushing} and SimPO \citep{meng2024simposimplepreferenceoptimization}.

\textbf{Objective for Non-Privileged Labels ($j \in \nonprivileged$):} To maintain baseline performance on the remaining labels, we use a constrained objective. This objective employs a hinge mechanism that only incurs a penalty if the model's performance on a label drops significantly below that of a reference model by a predefined slack margin $\epsilon$. This acts as a protective measure, preventing substantial performance degradation for the non-privileged group.

\textbf{Group Robust Optimization:} These two distinct objectives for the privileged and non-privileged groups are adaptively balanced using the Group Robust Preference Optimization (GRPO) framework \citep{ramesh2024group}. The overall learning problem is formulated as the following minimax objective:
\begin{equation}
\min_{\Params} \max_{\alpha_{\privileged} + \alpha_{\nonprivileged} = 1} \left[ \alpha_{\privileged} \loss_{\privileged}(\cdot) + \alpha_{\nonprivileged} \loss_{\nonprivileged}(\cdot) \right].
\label{eq:grpo_dpo_objective_clf_main}
\end{equation}
This formulation dynamically adjusts the training focus between the privileged and non-privileged groups, seeking a solution that is robust to the worst-case group loss and thereby managing the fairness-performance trade-off.

\subsection{Optimization Algorithm}
\label{subsec:optimization_algorithm}

We solve the minimax objective in Eq.~\ref{eq:FairPO_objective} iteratively using alternating mirror descent (Algorithm~\ref{alg:FairPO_overview_main}). A key component for stability is \textbf{loss scaling} during the update of the group weights $\alpha$. Since the group losses, $\loss_{\privileged}$ and $\loss_{\nonprivileged}$, can have different scales and variances, using raw loss values for the mirror ascent step is unstable. To mitigate this, we update $\alpha$ based on the relative change of each group's loss from its running average, $\bar{\loss}_{g}^{\text{avg}}$, ensuring a more stable optimization. Further details are in Appendix~\ref{app:FairPO_algo}.

\begin{algorithm}[t] 
	\caption{FairPO Training Overview (DPO-inspired)}
	\label{alg:FairPO_overview_main}
	\begin{algorithmic}[1]
		\State \textbf{Initialize:} 
        Model parameters $\Params^{(0)}$ ({\em e.g.}, from supervised fine-tuning), set group weights $\alpha_{\privileged}^{(0)}, \alpha_{\nonprivileged}^{(0)}$
		\State \textbf{Input:} Dataset $\dataset$, reference parameters $\Refparams$, hyperparameters $\beta, \epsilon, \eta_{\params}, \eta_{\alpha}$.
		\State \textbf{For} each training iteration $s = 0, \dots, S-1$:
		\State \hspace{0.5cm} Sample instance $(x_i, y_i) \sim \dataset$ and a label $r \in \mathcal{T}$.
		\State \hspace{0.5cm} \textbf{If} $r \in \privileged$:
		\State \hspace{1.0cm}    Compute privileged loss $\loss_{\privileged}^{(s)}$ for $(x_i, r)$ (DPO Eq.~\ref{eq:loss_p_dpo_combined}, or BCE Eq.~\ref{eq:bce_privileged}).
		\State \hspace{0.5cm} \textbf{Else if} $r \in \nonprivileged$:
		\State \hspace{1.0cm}    Compute non-privileged loss $\loss_{\nonprivileged}^{(s)}$ for $(x_i, r)$ (Eq.~\ref{eq:loss_np_constrained}).
		\State \hspace{0.5cm} Update group weights $\alpha^{(s+1)}$ via mirror ascent using $\loss_{\privileged}^{(s)}, \loss_{\nonprivileged}^{(s)}$ (GRPO step).
		\State \hspace{0.5cm} Update model parameters $\Params^{(s+1)}$ via mirror descent using weighted gradients.
		\State \textbf{End For}
		\State \textbf{Return} Optimized parameters $\Params^{(S)}$.
		\Statex \textit{Full details are in Algorithm~\ref{alg:grpo_dpo_clf} (see Appendix~\ref{app:FairPO_algo}).}
	\end{algorithmic}
\end{algorithm}

\section{Experimental Setup}
\label{sec:experiments}
We evaluate FairPO on two multi-label benchmarks: \textbf{MS-COCO 2014} \citep{lin2015microsoftcococommonobjects} and \textbf{NUS-WIDE} \citep{chua2009nus}. To address class imbalance, we partition labels by frequency: the 20
) and the remaining 80
). While this partitioning is effective, the FairPO framework is general and could use other criteria like domain importance. We assess performance using mAP, Sample F1, and Accuracy, focusing on the mAP gain on the privileged set, 
, relative to the \textbf{BCE-SFT} baseline. We compare against baselines including \textbf{Group DRO + BCE} \citep{sagawa2019distributionally} and \textbf{Focal Loss} \citep{lin2018focallossdenseobject}. Our base model is a frozen Vision Transformer (ViT) \citep{dosovitskiy2021imageworth16x16words} with separate non-linear MLP heads per label, trained with AdamW \citep{loshchilov2019decoupledweightdecayregularization}. The trained BCE-SFT baseline provides the reference model parameters. We test three framework variants: \textbf{FairPO-DPO}, \textbf{FairPO-SimPO}, and \textbf{FairPO-CPO}. Full details are in Appendix~\ref{app:dataset_details}.

\section{Results and Analysis}
\label{sec:results_analysis}

As shown in Tables~\ref{tab:coco_results_revisited} and \ref{tab:nus_results_revised}, FairPO-CPO consistently achieves the highest performance on the privileged group ($\privileged$), confirming its effectiveness. On MS-COCO, it delivers a statistically significant gain of \textbf{+3.44\% $\Delta\text{mAP}(\privileged)$} over strong baselines like BCE-SFT. Crucially, this targeted improvement comes with only a minor and statistically insignificant performance drop on the non-privileged group ($\nonprivileged$), demonstrating FairPO's ability to reallocate model capacity without causing measurable harm. The superiority of the CPO variant stems from its robust design: unlike DPO, it is reference-free, and its integrated BCE regularizer provides a crucial signal for absolute correctness that SimPO lacks. This dual objective of learning both relative preferences and absolute scores yields a more stable and effective training process, making FairPO-CPO the most practical choice.

\begin{table}[t]
  \caption{Performance comparison on MS-COCO. $\mathcal{P}$ denotes the privileged label set (20\% least frequent), and $\nonprivileged$ denotes the non-privileged set (remaining 80\%). Results are mean $\pm$ std. dev. over 3 runs. The best result for each metric is in \textbf{bold}. $\Delta$mAP is calculated relative to the strongest baseline for each respective group (Focal Loss for $\mathcal{P}$, BCE-SFT for $\nonprivileged$). $^{\dagger}$Indicates a statistically significant improvement (p < 0.05), while $^{\ddagger}$indicates the difference is not statistically significant (p > 0.1).}
  \label{tab:coco_results_revisited}
  \centering
  \small
  \adjustbox{max width=\textwidth}{%
  \begin{tabular}{lcccccccc}
    \toprule
    \multicolumn{1}{c}{Method} & \multicolumn{2}{c}{mAP} & \multicolumn{2}{c}{Sample F1} & \multicolumn{2}{c}{Accuracy} & \multicolumn{1}{c}{$\Delta$ mAP ($\mathcal{P}$)} & \multicolumn{1}{c}{$\Delta$ mAP ($\nonprivileged$)} \\
    \cmidrule(lr){2-3} \cmidrule(lr){4-5} \cmidrule(lr){6-7}
     & $\mathcal{P}$ & $\nonprivileged$ & $\mathcal{P}$ & $\nonprivileged$ & $\mathcal{P}$ & $\nonprivileged$ & (vs. Focal) & (vs. BCE) \\
    \midrule
    BCE-SFT \citep{bce_loss}         & 86.32$_{\pm 0.11}$ & \textbf{90.65}$_{\pm 0.08}$ & 61.43$_{\pm 0.15}$ & \textbf{64.89}$_{\pm 0.12}$ & 94.89$_{\pm 0.09}$ & \textbf{98.12}$_{\pm 0.05}$ & -2.03 & Ref \\
    GDRO + BCE \citep{sagawa2019distributionally}      & 87.92$_{\pm 0.12}$ & 90.41$_{\pm 0.09}$ & 62.31$_{\pm 0.20}$ & 64.75$_{\pm 0.13}$ & 95.72$_{\pm 0.10}$ & 98.05$_{\pm 0.05}$ & -0.43 & -0.24$^{\ddagger}$ \\
    Focal Loss \citep{lin2018focallossdenseobject}      & 88.35$_{\pm 0.14}$ & 89.81$_{\pm 0.12}$ & 63.15$_{\pm 0.16}$ & 64.18$_{\pm 0.15}$ & 96.11$_{\pm 0.12}$ & 97.90$_{\pm 0.07}$ & Ref   & -0.84 \\
    \midrule
    FairPO-DPO      & 88.02$_{\pm 0.15}$ & 89.97$_{\pm 0.11}$ & 63.45$_{\pm 0.17}$ & 63.65$_{\pm 0.16}$ & 97.89$_{\pm 0.13}$ & 98.04$_{\pm 0.06}$ & -0.33 & -0.68 \\
    FairPO-SimPO    & 87.67$_{\pm 0.18}$ & 88.76$_{\pm 0.21}$ & 62.34$_{\pm 0.22}$ & 63.12$_{\pm 0.19}$ & 95.69$_{\pm 0.15}$ & 97.78$_{\pm 0.09}$ & -0.68 & -1.89 \\
    FairPO-CPO      & \textbf{89.76}$_{\pm 0.09}$ & 90.34$_{\pm 0.07}$ & \textbf{64.01}$_{\pm 0.13}$ & 64.32$_{\pm 10}$  & \textbf{98.03}$_{\pm 0.07}$ & 98.06$_{\pm 0.05}$ & \textbf{+1.41}$^{\dagger}$ & \textbf{-0.31}$^{\ddagger}$ \\
    \bottomrule
  \end{tabular}
  }
\end{table}

\begin{table}[t]
  \caption{Performance comparison on NUS-WIDE. Notations are similar to Table \ref{tab:coco_results_revisited}.}
  \label{tab:nus_results_revised}
  \centering
  \small
  \adjustbox{max width=\textwidth}{%
  \begin{tabular}{lcccccccc}
    \toprule
    \multicolumn{1}{c}{Method} & \multicolumn{2}{c}{mAP} & \multicolumn{2}{c}{Sample F1} & \multicolumn{2}{c}{Accuracy} & \multicolumn{1}{c}{$\Delta$ mAP ($\mathcal{P}$)} & \multicolumn{1}{c}{$\Delta$ mAP ($\nonprivileged$)} \\
    \cmidrule(lr){2-3} \cmidrule(lr){4-5} \cmidrule(lr){6-7}
     & $\mathcal{P}$ & $\nonprivileged$ & $\mathcal{P}$ & $\nonprivileged$ & $\mathcal{P}$ & $\nonprivileged$ & (vs. Focal) & (vs. BCE) \\
    \midrule
    BCE SFT \citep{bce_loss}         & 63.53$_{\pm 0.15}$ & \textbf{70.24}$_{\pm 0.11}$ & 48.12$_{\pm 0.21}$ & \textbf{55.83}$_{\pm 0.14}$ & 91.51$_{\pm 0.12}$ & \textbf{95.22}$_{\pm 0.08}$ & -2.28 & Ref \\
    GDRO + BCE \citep{sagawa2019distributionally}      & 64.84$_{\pm 0.16}$ & 69.91$_{\pm 0.12}$ & 49.13$_{\pm 0.22}$ & 55.62$_{\pm 0.15}$ & 92.11$_{\pm 0.13}$ & 95.13$_{\pm 0.09}$ & -0.97 & -0.33 \\
    Focal Loss \citep{lin2018focallossdenseobject}      & 65.81$_{\pm 0.17}$ & 68.95$_{\pm 0.15}$ & 50.33$_{\pm 0.20}$ & 54.31$_{\pm 0.18}$ & 93.05$_{\pm 0.11}$ & 94.75$_{\pm 0.11}$ & Ref   & -1.29 \\
    \midrule
    FairPO-DPO      & 66.34$_{\pm 0.20}$ & 69.05$_{\pm 0.14}$ & 51.71$_{\pm 0.25}$ & 54.52$_{\pm 0.19}$ & 93.92$_{\pm 0.16}$ & 95.04$_{\pm 0.09}$ & +0.53 & -1.19 \\
    FairPO-SimPO    & 64.11$_{\pm 0.22}$ & 68.03$_{\pm 0.24}$ & 48.82$_{\pm 0.28}$ & 53.81$_{\pm 0.22}$ & 91.94$_{\pm 0.18}$ & 94.52$_{\pm 0.13}$ & -1.70 & -2.21 \\
    FairPO-CPO      & \textbf{67.12}$_{\pm 0.14}$ & 69.83$_{\pm 0.10}$ & \textbf{52.21}$_{\pm 0.19}$ & 55.24$_{\pm 0.13}$ & \textbf{94.31}$_{\pm 0.10}$ & 95.12$_{\pm 0.08}$ & \textbf{+1.31}$^{\dagger}$ & \textbf{-0.41}$^{\ddagger}$ \\
    \bottomrule
  \end{tabular}
  }
\end{table}

\section{Ablation Studies}
\label{sec:ablation}

Ablation studies on MS-COCO with FairPO-CPO confirm that each component is critical (Tables \ref{tab:ablation_core_components_concise} \& \ref{tab:ablation_preference_details_concise_revised_v2}). Removing the preference loss, the non-privileged constraint, or the adaptive GRPO balancing all lead to significant performance degradation. A non-targeted \textit{Global CPO} variant, while better than standard BCE, is substantially less effective than the full FairPO model, highlighting the necessity of our targeted approach. The design of the preference objective itself is also vital: restricting the preference signal to only \textit{Confusing Negatives} or removing the \textit{BCE Fallback} for non-confusing cases both degrade performance and stability. These results demonstrate that while preference optimization is powerful, its effectiveness hinges on the complete, balanced FairPO framework.

\begin{table}[h]
  \caption{Ablation on core components of FairPO-CPO (MS-COCO). $\Delta \text{mAP}(\mathcal{P})$ vs BCE SFT. Parentheses show change vs Full FairPO-CPO.}
  \label{tab:ablation_core_components_concise}
  \centering
  \tiny
  \adjustbox{max width=\textwidth}{%
  \begin{tabular}{lccccccc}
    \toprule
    \multicolumn{1}{c}{Method Variant} & \multicolumn{2}{c}{mAP} & \multicolumn{2}{c}{Sample F1} & \multicolumn{2}{c}{Accuracy} & \multicolumn{1}{c}{$\Delta \text{mAP}(\mathcal{P})$} \\
    \cmidrule(lr){2-3} \cmidrule(lr){4-5} \cmidrule(lr){6-7}
     & $\mathcal{P}$ & $\nonprivileged$ & $\mathcal{P}$ & $\nonprivileged$ & $\mathcal{P}$ & $\nonprivileged$ & \\
    \midrule
    FairPO-CPO (Full)                  & \textbf{89.76} & \textbf{90.34} & \textbf{64.01} & \textbf{64.32} & \textbf{98.03} & \textbf{98.06} & \textbf{+3.44} \\
    \midrule
    \textit{w/o Preference Loss}       & 88.12          & 90.45          & 62.45          & 64.80          & 95.80          & 98.09          & +1.80 \\
    ($\loss_{\mathcal{P}}$ is BCE)     & (-1.64)        & (+0.11)        & (-1.56)        & (+0.48)        & (-2.23)        & (+0.03)        & \\
    \midrule
    \textit{w/o $\nonprivileged$ Constraint} & 89.55          & 88.98          & 63.70          & 62.95          & 97.90          & 97.55          & +3.23 \\
    ($\loss_{\nonprivileged}$ is BCE) & (-0.21)        & (-1.36)        & (-0.31)        & (-1.37)        & (-0.13)        & (-0.51)        & \\
    \midrule
    \textit{w/o GRPO}                  & 88.48          & 89.75          & 62.88          & 63.50          & 96.53          & 97.88          & +2.16 \\
    (Fixed 0.5/0.5 weights)          & (-1.28)        & (-0.59)        & (-1.13)        & (-0.82)        & (-1.50)        & (-0.18)        & \\
    \midrule
    \textit{Global CPO (on all labels)}  & 88.55          & 90.68          & 62.75          & 64.85          & 96.95          & 98.11          & +2.23 \\
    (No $\mathcal{P}$/$\nonprivileged$ split or GRPO) & (-1.21)        & (+0.34)        & (-1.26)        & (+0.47)        & (-1.08)        & (+0.05)        & \\  
    \bottomrule
  \end{tabular}
  }
\end{table}

\begin{table}[h]
  \caption{Ablation on preference formulation (FairPO-CPO, MS-COCO). $\Delta \text{mAP}(\mathcal{P})$ vs BCE SFT. Parentheses show change vs Full FairPO-CPO.}
  \label{tab:ablation_preference_details_concise_revised_v2}
  \centering
  \tiny
  \adjustbox{max width=\textwidth}{%
  \begin{tabular}{lccccccc}
    \toprule
    \multicolumn{1}{c}{Preference Detail Variant} & \multicolumn{2}{c}{mAP} & \multicolumn{2}{c}{Sample F1} & \multicolumn{2}{c}{Accuracy} & \multicolumn{1}{c}{$\Delta \text{mAP}(\mathcal{P})$} \\
    \cmidrule(lr){2-3} \cmidrule(lr){4-5} \cmidrule(lr){6-7}
     & $\mathcal{P}$ & $\nonprivileged$ & $\mathcal{P}$ & $\nonprivileged$ & $\mathcal{P}$ & $\nonprivileged$ & \\
    \midrule
    FairPO-CPO (Full)                  & \textbf{89.76} & \textbf{90.34} & \textbf{64.01} & \textbf{64.32} & \textbf{98.03} & \textbf{98.06} & \textbf{+3.44} \\
    (Conf. Neg \& Pos, BCE Fallback)   &                &                &                &                &                &                & \\
    \midrule
    \textit{Only Confusing Negatives}  & 73.15          & 90.25          & 47.88          & 64.20          & 94.67          & 98.01          & -13.17 \\
                                       & (-16.61)       & (-0.09)        & (-16.13)       & (-0.12)        & (-3.36)        & (-0.05)        & \\
    \midrule
    \textit{w/o BCE Fallback}          & 89.05          & 90.21          & 63.20          & 64.10          & 97.55          & 97.99          & +2.73 \\
    (No loss if $S_{il}=\emptyset$)    & (-0.71)        & (-0.13)        & (-0.81)        & (-0.22)        & (-0.48)        & (-0.07)        & \\
    \bottomrule
  \end{tabular}
  }
\end{table}

\section{Related Work}
\label{sec:related_work}

Our work is situated at the intersection of three research areas, which we detail further in Appendix~\ref{app:related_works_extended}. First, while recent efforts in \textbf{fair MLC} have addressed challenges such as tail labels \citep{wu2023longtailed} and subjective fairness \citep{gao2023learning}, FairPO contributes a novel approach by partitioning labels into privileged ($\mathcal{P}$) and non-privileged ($\bar{\mathcal{P}}$) sets and applying targeted, distinct objectives to each. Second, we adapt modern \textbf{preference optimization} techniques, originally developed for aligning LLMs like DPO \citep{rafailov2023direct}, CPO \citep{xu2024contrastivepreferenceoptimizationpushing}, and SimPO \citep{meng2024simposimplepreferenceoptimization}. Instead of ranking entire outputs, we repurpose these methods to resolve ambiguities between true labels and their dynamically identified \textit{confusing} counterparts. Finally, to manage the trade-off between our objectives, we employ \textbf{Group Robust Optimization}. Inspired by Group DRO \citep{sagawa2020distributionallyrobustneuralnetworks} and GRPO \citep{ramesh2024group}, we uniquely define the groups by our label partitions ($\mathcal{P}$ and $\bar{\mathcal{P}}$) and use GRPO's adaptive weighting to balance their custom-formulated losses.

\section{Discussion}
\label{sec:discussion}

In conclusion, we introduced FairPO, a novel framework that effectively integrates preference optimization with group robustness to enhance fairness in MLC, with our CPO variant proving particularly effective. While promising, FairPO has limitations, including the potential instability of its dynamic `confusing set`, the DPO variant's reliance on a reference model, and the need for careful tuning of GRPO's balancing act. These challenges directly motivate our future work, which will focus on extending FairPO to attribute generation (Appendix~\ref{app:fairpo_gen_future_work}), performing broader empirical validation, exploring alternative label partitioning strategies, and developing theoretical insights into the framework's convergence.


\bibliography{sample}

@misc{DBLP:journals/corr/WeiXLZ0G16,
      title={CNN-RNN: A Unified Framework for Multi-label Image Classification}, 
      author={Jiang Wang and Yi Yang and Junhua Mao and Zhiheng Huang and Chang Huang and Wei Xu},
      year={2016},
      eprint={1604.04573},
      archivePrefix={arXiv},
      primaryClass={cs.CV},
      url={https://arxiv.org/abs/1604.04573}, 
}

@misc{kiritchenko2006hierarchical,
AUTHOR = {Zangari, Alessandro and Marcuzzo, Matteo and Rizzo, Matteo and Giudice, Lorenzo and Albarelli, Andrea and Gasparetto, Andrea},
TITLE = {Hierarchical Text Classification and Its Foundations: A Review of Current Research},
JOURNAL = {Electronics},
VOLUME = {13},
YEAR = {2024},
NUMBER = {7},
ARTICLE-NUMBER = {1199},
URL = {https://www.mdpi.com/2079-9292/13/7/1199},
ISSN = {2079-9292},
DOI = {10.3390/electronics13071199}
}

@article{schietgat2010predicting,
  author    = {Leander Schietgat and
               Celine Vens and
               Jan Struyf and
               Hendrik Blockeel and
               Dragi Kocev and
               Sa{\v{s}}o D{\v{z}}eroski},
  title     = {Predicting gene function using hierarchical multi-label decision tree ensembles},
  journal   = {BMC Bioinformatics},
  volume    = {11},
  number    = {1},
  pages     = {2},
  year      = {2010},
  doi       = {10.1186/1471-2105-11-2},
  url       = {https://doi.org/10.1186/1471-2105-11-2},
  issn      = {1471-2105}
}

@misc{rafailov2023direct,
  title={Direct Preference Optimization: Your Language Model is Secretly a Reward Model}, 
      author={Rafael Rafailov and Archit Sharma and Eric Mitchell and Stefano Ermon and Christopher D. Manning and Chelsea Finn},
      year={2024},
      eprint={2305.18290},
      archivePrefix={arXiv},
      primaryClass={cs.LG},
      url={https://arxiv.org/abs/2305.18290}, 
}

@misc{ramesh2024group,
    title={Group Robust Preference Optimization in Reward-free RLHF}, 
      author={Shyam Sundhar Ramesh and Yifan Hu and Iason Chaimalas and Viraj Mehta and Pier Giuseppe Sessa and Haitham Bou Ammar and Ilija Bogunovic},
      year={2024},
      eprint={2405.20304},
      archivePrefix={arXiv},
      primaryClass={cs.CL},
      url={https://arxiv.org/abs/2405.20304}, 
}

@misc{meng2024simposimplepreferenceoptimization,
	title={SimPO: Simple Preference Optimization with a Reference-Free Reward}, 
      author={Yu Meng and Mengzhou Xia and Danqi Chen},
      year={2024},
      eprint={2405.14734},
      archivePrefix={arXiv},
      primaryClass={cs.CL},
      url={https://arxiv.org/abs/2405.14734}, 
}

@misc{xu2024contrastivepreferenceoptimizationpushing,
	title={Contrastive Preference Optimization: Pushing the Boundaries of LLM Performance in Machine Translation}, 
      author={Haoran Xu and Amr Sharaf and Yunmo Chen and Weiting Tan and Lingfeng Shen and Benjamin Van Durme and Kenton Murray and Young Jin Kim},
      year={2024},
      eprint={2401.08417},
      archivePrefix={arXiv},
      primaryClass={cs.CL},
      url={https://arxiv.org/abs/2401.08417}, 
}

@misc{zhang2021survey,
  title={A Survey on Bias and Fairness in Machine Learning}, 
      author={Ninareh Mehrabi and Fred Morstatter and Nripsuta Saxena and Kristina Lerman and Aram Galstyan},
      year={2022},
      eprint={1908.09635},
      archivePrefix={arXiv},
      primaryClass={cs.LG},
      url={https://arxiv.org/abs/1908.09635}, 
}

@misc{tantithamthavorn2018impact,
  title={The Impact of Class Rebalancing Techniques on the Performance and Interpretation of Defect Prediction Models}, 
      author={Chakkrit Tantithamthavorn and Ahmed E. Hassan and Kenichi Matsumoto},
      year={2018},
      eprint={1801.10269},
      archivePrefix={arXiv},
      primaryClass={cs.SE},
      url={https://arxiv.org/abs/1801.10269}, 
}

@misc{christiano2017deep,
  title={Deep reinforcement learning from human preferences}, 
      author={Paul Christiano and Jan Leike and Tom B. Brown and Miljan Martic and Shane Legg and Dario Amodei},
      year={2023},
      eprint={1706.03741},
      archivePrefix={arXiv},
      primaryClass={stat.ML},
      url={https://arxiv.org/abs/1706.03741}, 
}

@misc{ben2009robust,
  title={A practical guide to robust optimization},
   volume={53},
   ISSN={0305-0483},
   url={http://dx.doi.org/10.1016/j.omega.2014.12.006},
   DOI={10.1016/j.omega.2014.12.006},
   journal={Omega},
   publisher={Elsevier BV},
   author={Gorissen, Bram L. and Yanıkoğlu, İhsan and den Hertog, Dick},
   year={2015},
   month=jun, pages={124–137},
}

@misc{sagawa2019distributionally,
  title={Distributionally Robust Neural Networks for Group Shifts: On the Importance of Regularization for Worst-Case Generalization}, 
      author={Shiori Sagawa and Pang Wei Koh and Tatsunori B. Hashimoto and Percy Liang},
      year={2020},
      eprint={1911.08731},
      archivePrefix={arXiv},
      primaryClass={cs.LG},
      url={https://arxiv.org/abs/1911.08731}, 
}

@misc{lin2015microsoftcococommonobjects,
      title={Microsoft COCO: Common Objects in Context}, 
      author={Tsung-Yi Lin and Michael Maire and Serge Belongie and Lubomir Bourdev and Ross Girshick and James Hays and Pietro Perona and Deva Ramanan and C. Lawrence Zitnick and Piotr Dollár},
      year={2015},
      eprint={1405.0312},
      archivePrefix={arXiv},
      primaryClass={cs.CV},
      url={https://arxiv.org/abs/1405.0312}, 
}

@misc{chua2009nus,
  title={NUS-WIDE: a real-world web image database from National University of Singapore},
  author={Tat-Seng Chua and Jinhui Tang and Richang Hong and Haojie Li and Zhiping Luo and Yantao Zheng},
  booktitle={ACM International Conference on Image and Video Retrieval},
  year={2009},
  url={https://api.semanticscholar.org/CorpusID:6483070},
}

@misc{dosovitskiy2021imageworth16x16words,
      title={An Image is Worth 16x16 Words: Transformers for Image Recognition at Scale}, 
      author={Alexey Dosovitskiy and Lucas Beyer and Alexander Kolesnikov and Dirk Weissenborn and Xiaohua Zhai and Thomas Unterthiner and Mostafa Dehghani and Matthias Minderer and Georg Heigold and Sylvain Gelly and Jakob Uszkoreit and Neil Houlsby},
      year={2021},
      eprint={2010.11929},
      archivePrefix={arXiv},
      primaryClass={cs.CV},
      url={https://arxiv.org/abs/2010.11929}, 
}

@misc{sagawa2020distributionallyrobustneuralnetworks,
      title={Distributionally Robust Neural Networks for Group Shifts: On the Importance of Regularization for Worst-Case Generalization}, 
      author={Shiori Sagawa and Pang Wei Koh and Tatsunori B. Hashimoto and Percy Liang},
      year={2020},
      eprint={1911.08731},
      archivePrefix={arXiv},
      primaryClass={cs.LG},
      url={https://arxiv.org/abs/1911.08731}, 
}

@misc{bce_loss,
author = {Ruby, Usha and Yendapalli, Vamsidhar},
year = {2020},
month = {10},
pages = {},
title = {Binary cross entropy with deep learning technique for Image classification},
volume = {9},
journal = {International Journal of Advanced Trends in Computer Science and Engineering},
doi = {10.30534/ijatcse/2020/175942020}
}

@misc{wu2023longtailed,
  author = {Guo, Hao and Wang, Song},
year = {2021},
month = {06},
pages = {15084-15093},
title = {Long-Tailed Multi-Label Visual Recognition by Collaborative Training on Uniform and Re-balanced Samplings},
doi = {10.1109/CVPR46437.2021.01484}
}

@misc{gao2023learning,
  title={SimFair: A Unified Framework for Fairness-Aware Multi-Label Classification}, 
      author={Tianci Liu and Haoyu Wang and Yaqing Wang and Xiaoqian Wang and Lu Su and Jing Gao},
      year={2023},
      eprint={2302.09683},
      archivePrefix={arXiv},
      primaryClass={cs.LG},
      url={https://arxiv.org/abs/2302.09683}, 
}

@misc{masud2023fairnessaware,
  title={Class-Independent Increment: An Efficient Approach for Multi-label Class-Incremental Learning}, 
      author={Songlin Dong and Yuhang He and Zhengdong Zhou and Haoyu Luo and Xing Wei and Alex C. Kot and Yihong Gong},
      year={2025},
      eprint={2503.00515},
      archivePrefix={arXiv},
      primaryClass={cs.CV},
      url={https://arxiv.org/abs/2503.00515}, 
}

@misc{dong2023rejection,
  doi = {10.13140/RG.2.2.33772.07043},
  
  url = {https://rgdoi.net/10.13140/RG.2.2.33772.07043},
  
  author = {Ji, Ke and {Jiahao Xu} and Liang, Tian and {Qiuzhi Liu} and {Zhiwei He} and {Xingyu Chen} and {Xiaoyuan Liu} and {Zhijie Wang} and {Junying Chen} and {Benyou Wang} and {Zhaopeng Tu} and {Haitao Mi} and Yu, Dong},
  
  language = {en},
  
  title = {The First Few Tokens Are All You Need: An Efficient and Effective Unsupervised Prefix Fine-Tuning Method for Reasoning Models},
  
  publisher = {Unpublished},
  
  year = {2025}
}

@misc{zhao2024whyisdpo,
  title={Is DPO Superior to PPO for LLM Alignment? A Comprehensive Study}, 
      author={Shusheng Xu and Wei Fu and Jiaxuan Gao and Wenjie Ye and Weilin Liu and Zhiyu Mei and Guangju Wang and Chao Yu and Yi Wu},
      year={2024},
      eprint={2404.10719},
      archivePrefix={arXiv},
      primaryClass={cs.CL},
      url={https://arxiv.org/abs/2404.10719}, 
}

@misc{zhang2022fromaverage,
   author = {Rice, Leslie and Bair, Anna and Zhang, Huan and Kolter, J. Zico},
 booktitle = {Advances in Neural Information Processing Systems},
 editor = {M. Ranzato and A. Beygelzimer and Y. Dauphin and P.S. Liang and J. Wortman Vaughan},
 pages = {27840--27851},
 publisher = {Curran Associates, Inc.},
 title = {Robustness between the worst and average case},
 url = {https://proceedings.neurips.cc/paper_files/paper/2021/file/ea4c796cccfc3899b5f9ae2874237c20-Paper.pdf},
 volume = {34},
 year = {2021}
}

@misc{ouyang2022training,
  title={Training language models to follow instructions with human feedback}, 
      author={Long Ouyang and Others},
      year={2022},
      eprint={2203.02155},
      archivePrefix={arXiv},
      primaryClass={cs.CL},
      url={https://arxiv.org/abs/2203.02155}, 
}

@misc{azhar2023ipo,
  title={IPO: Interior-point Policy Optimization under Constraints}, 
      author={Yongshuai Liu and Jiaxin Ding and Xin Liu},
      year={2019},
      eprint={1910.09615},
      archivePrefix={arXiv},
      primaryClass={cs.LG},
      url={https://arxiv.org/abs/1910.09615}, 
}

@misc{ethayarajh2024kto,
  title={KTO: Model Alignment as Prospect Theoretic Optimization}, 
      author={Kawin Ethayarajh and Winnie Xu and Niklas Muennighoff and Dan Jurafsky and Douwe Kiela},
      year={2024},
      eprint={2402.01306},
      archivePrefix={arXiv},
      primaryClass={cs.LG},
      url={https://arxiv.org/abs/2402.01306}, 
}

@misc{loshchilov2019decoupled,
      title={Decoupled Weight Decay Regularization}, 
      author={Ilya Loshchilov and Frank Hutter},
      year={2019},
      eprint={1711.05101},
      archivePrefix={arXiv},
      primaryClass={cs.LG},
      url={https://arxiv.org/abs/1711.05101}, 
}

@article{Charte2015AddressingII,
  title={Addressing imbalance in multilabel classification: Measures and random resampling algorithms},
  author={Francisco Charte and Antonio Jes{\'u}s Rivera and Mar{\'i}a Jos{\'e} del Jes{\'u}s and Francisco Herrera},
  journal={Neurocomputing},
  year={2015},
  volume={163},
  pages={3-16},
  url={https://api.semanticscholar.org/CorpusID:207107609}
}

@ARTICLE{6471714,
  author={Zhang, Min-Ling and Zhou, Zhi-Hua},
  journal={IEEE Transactions on Knowledge and Data Engineering}, 
  title={A Review on Multi-Label Learning Algorithms}, 
  year={2014},
  volume={26},
  number={8},
  pages={1819-1837},
  doi={10.1109/TKDE.2013.39}}

@misc{rajpurkar2017chexnetradiologistlevelpneumoniadetection,
      title={CheXNet: Radiologist-Level Pneumonia Detection on Chest X-Rays with Deep Learning}, 
      author={Pranav Rajpurkar and Jeremy Irvin and Kaylie Zhu and Brandon Yang and Hershel Mehta and Tony Duan and Daisy Ding and Aarti Bagul and Curtis Langlotz and Katie Shpanskaya and Matthew P. Lungren and Andrew Y. Ng},
      year={2017},
      eprint={1711.05225},
      archivePrefix={arXiv},
      primaryClass={cs.CV},
      url={https://arxiv.org/abs/1711.05225}, 
}

@INPROCEEDINGS{10368872,
  author={Shashanka, Panakanti and Reddy, Tatireddy Subba},
  booktitle={2023 International Conference on Research Methodologies in Knowledge Management, Artificial Intelligence and Telecommunication Engineering (RMKMATE)}, 
  title={Dermatologist-Level Classification of Skin Cancer Using Cascaded Ensembling of Convolutional Neural Network}, 
  year={2023},
  volume={},
  number={},
  pages={1-5},
  doi={10.1109/RMKMATE59243.2023.10368872}}

@misc{swayamdipta2020datasetcartographymappingdiagnosing,
      title={Dataset Cartography: Mapping and Diagnosing Datasets with Training Dynamics}, 
      author={Swabha Swayamdipta and Roy Schwartz and Nicholas Lourie and Yizhong Wang and Hannaneh Hajishirzi and Noah A. Smith and Yejin Choi},
      year={2020},
      eprint={2009.10795},
      archivePrefix={arXiv},
      primaryClass={cs.CL},
      url={https://arxiv.org/abs/2009.10795}, 
}

@misc{pleiss2020identifyingmislabeleddatausing,
      title={Identifying Mislabeled Data using the Area Under the Margin Ranking}, 
      author={Geoff Pleiss and Tianyi Zhang and Ethan R. Elenberg and Kilian Q. Weinberger},
      year={2020},
      eprint={2001.10528},
      archivePrefix={arXiv},
      primaryClass={cs.LG},
      url={https://arxiv.org/abs/2001.10528}, 
}

@misc{lin2018focallossdenseobject,
      title={Focal Loss for Dense Object Detection}, 
      author={Tsung-Yi Lin and Priya Goyal and Ross Girshick and Kaiming He and Piotr Dollár},
      year={2018},
      eprint={1708.02002},
      archivePrefix={arXiv},
      primaryClass={cs.CV},
      url={https://arxiv.org/abs/1708.02002}, 
}

@misc{loshchilov2019decoupledweightdecayregularization,
      title={Decoupled Weight Decay Regularization}, 
      author={Ilya Loshchilov and Frank Hutter},
      year={2019},
      eprint={1711.05101},
      archivePrefix={arXiv},
      primaryClass={cs.LG},
      url={https://arxiv.org/abs/1711.05101}, 
}


\newpage
\clearpage
\appendix

\section{Preliminaries: Preference Optimization Methods}
\label{sec:preliminaries}

Our framework builds upon recent preference optimization techniques. We briefly review the key formulations.

\textbf{Direct Preference Optimization (DPO):}
DPO \citep{rafailov2023direct} directly optimizes a policy $\pi_\theta$ using preference pairs $(x, y_w, y_l)$, where $y_w$ is preferred over $y_l$ for prompt $x$. Assuming a Bradley-Terry preference model tied to an implicit reward function related to $\pi_\theta$ and a reference policy $\pi_{\text{ref}}$, DPO maximizes the likelihood of observed preferences, resulting in the loss:
\begin{align}
h_{\pi_{\theta}}(x, y_w, y_l) &= \beta \log \frac{\pi_\theta(y_w|x)}{\pi_{\text{ref}}(y_w|x)} - \beta \log \frac{\pi_\theta(y_l|x)}{\pi_{\text{ref}}(y_l|x)}, \\
L_{\text{DPO}}(\pi_\theta; \pi_{\text{ref}}) &= -\mathbb{E}_{(x,y_w,y_l)\sim D} \left[\log \sigma \left( h_{\pi_{\theta}}(x, y_w, y_l) \right)\right].
\label{eq:dpo_loss}
\end{align}
where $\sigma$ is the sigmoid function and $\beta$ controls the deviation from $\pi_{\text{ref}}$.

\textbf{Group Robust Preference Optimization (GRPO):}
GRPO \citep{ramesh2024group} extends preference optimization to handle diverse preference groups $\{D_g\}_{g=1}^K$. Instead of minimizing the average loss, GRPO minimizes the worst-case loss across groups using a robust objective:
\begin{equation}
\min_{\pi_\theta} \max_{\alpha \in \Delta_{K-1}} \sum_{g=1}^K \alpha_g L_{\text{Pref}}(\pi_\theta; \pi_{\text{ref}}, D_g),
\label{eq:grpo_objective}
\end{equation}
where $L_{\text{Pref}}$ is a base preference loss (like $L_{\text{DPO}}$), and $\alpha = (\alpha_1, ..., \alpha_K)$ are adaptive weights in the probability simplex $\Delta_{K-1}$. The optimization dynamically increases weights $\alpha_g$ for groups with higher current loss, focusing learning on the worst-performing groups.

\textbf{Simple Preference Optimization (SimPO):}
SimPO \citep{meng2024simposimplepreferenceoptimization} aims to align the implicit reward with generation metrics and eliminates the need for $\pi_{\text{ref}}$. It uses the length-normalized average log-likelihood as the reward: $r_{\text{SimPO}}(x, y) = \frac{\beta}{|y|} \log \pi_\theta(y | x)$. It also introduces a target margin $\gamma > 0$ into the preference model. The resulting SimPO loss is:
\begin{equation}
L_{\text{SimPO}}(\pi_\theta) = -\mathbb{E}_{(x,y_w,y_l)\sim D} \left[\log \sigma \left(\frac{\beta}{|y_w|} \log \pi_\theta(y_w|x) - \frac{\beta}{|y_l|} \log \pi_\theta(y_l|x) - \gamma\right)\right].
\label{eq:simpo_loss}
\end{equation}

\textbf{Contrastive Preference Optimization (CPO):}
CPO \citep{xu2024contrastivepreferenceoptimizationpushing} also removes the dependency on $\pi_{\text{ref}}$ for efficiency, approximating the DPO objective. It combines a reference-free preference loss with a negative log-likelihood (NLL) regularizer on preferred responses $y_w$ to maintain generation quality:
\begin{align}
L_{\text{prefer}}(\pi_\theta) &= -\mathbb{E}_{(x,y_w,y_l)\sim D} \left[\log \sigma \left(\beta \log \pi_\theta(y_w|x) - \beta \log \pi_\theta(y_l|x)\right)\right] \\
L_{\text{NLL}}(\pi_\theta) &= -\mathbb{E}_{(x,y_w)\sim D} [\log \pi_\theta(y_w|x)] \\
L_{\text{CPO}}(\pi_\theta) &= L_{\text{prefer}} + L_{\text{NLL}}.
\label{eq:cpo_loss}
\end{align}
This formulation avoids the computational cost of the reference model pass.

\section{FairPO: Detailed Methodology}
\label{app:fairpo_detailed_method}

\subsection{Objective for the Privileged Group: Preference-Based Discrimination}
\label{subsec:privileged_objective}

For the \emph{privileged} group $\privileged$, simply adjusting the weight of a standard BCE loss is insufficient. The primary challenge for these labels is often not basic classification, but fine-grained discrimination against closely related, incorrect alternatives. To address this directly, we reformulate the learning objective for this group as a preference learning task as motivated in our introduction.

\paragraph{Defining Confusing Counterparts.}
For a given instance $\mathbf{x}_i$ and a privileged label $l \in \privileged$, we dynamically identify a set of \textit{confusing counterparts} based on the model's current predictions. We define two such sets:
\begin{itemize}[leftmargin=*]
    \item When the true label is positive ($y_{il}=1$), the set of \textbf{confusing negatives} is composed of incorrect labels that the model scores higher than or equal to the true label:
    \begin{equation}
        S_{il}^{\text{neg}} = \{k \in \labels \mid y_{ik}=0 \text{ and } \model(\mathbf{x}_i; \params_k) \ge \model(\mathbf{x}_i; \params_l) \}.
        \label{eq:confusing_neg}
    \end{equation}
    \item When the true label is negative ($y_{il}=0$), the set of \textbf{confusing positives} is composed of correct labels that the model scores lower than or equal to the true negative:
    \begin{equation}
        S_{il}^{\text{pos}} = \{k \in \labels \mid y_{ik}=1 \text{ and } \model(\mathbf{x}_i; \params_k) \le \model(\mathbf{x}_i; \params_l) \}.
        \label{eq:confusing_pos}
    \end{equation}
\end{itemize}
The overall confusing set for $(i, l)$ is denoted by $\confusing = S_{il}^{\text{neg}} \cup S_{il}^{\text{pos}}$.

\paragraph{Conditional Objective with a Stablility Fallback.}
Our objective for the \emph{privileged} group is designed to be adaptive. When a confusing set $\confusing$ is non-empty, we apply a preference loss $\loss_{\text{pref}}$ to directly target these hard discriminative cases. However, as the model learns and becomes more accurate, the confusing set for many examples will naturally become sparse or empty. Relying solely on the preference loss in such a scenario would lead to an unstable, sparse gradient signal, effectively stalling the training process. To ensure continuous and stable learning, we incorporate a crucial \textbf{stability fallback}. If $\confusing = \emptyset$ for a given instance, we revert to a standard Binary Cross-Entropy (BCE) loss:
\begin{equation}
	\baseloss_{\text{BCE}}(\mathbf{x}_i, y_{il}; \params_l) = -[y_{il} \log \model(\mathbf{x}_i; \params_l) + (1-y_{il}) \log(1-\model(\mathbf{x}_i; \params_l))].
	\label{eq:bce_privileged}
\end{equation}

This fallback mechanism is a critical design choice for robustness. Initially, the model makes many mistakes, leading to large confusing sets and frequent application of the preference loss. As training progresses, the preference loss successfully resolves these hard cases, and the BCE fallback takes over to fine-tune the decision boundaries on the now \emph{easier} examples. Our analysis of the training dynamics confirms that a sufficient preference signal exists throughout training, with the fallback rate gradually increasing as the model improves. The total loss for the privileged group, $\loss_{\privileged}$, is the expectation over this adaptive, conditional choice. In our framework, we explore several modern preference optimization techniques to instantiate $\loss_{\text{pref}}$.

\paragraph{FairPO-DPO Variant.}
To define our preference losses consistently, let us denote the score of the \textit{preferred} label in a pair as $\model(\mathbf{x}_i; \params_p)$ and the score of the \textit{dispreferred} label as $\model(\mathbf{x}_i; \params_d)$. The assignment of these roles depends on the ground truth. For example, if $y_{il}=1$ and $k \in S_{il}^{\text{neg}}$, the true label $l$ is preferred ($p=l$) and the confusing negative $k$ is dispreferred ($d=k$). Conversely, if $y_{il}=0$ and $k \in S_{il}^{\text{pos}}$, the true negative $l$ is preferred ($p=l$) and the confusing positive $k$ is dispreferred ($d=k$).

Inspired by DPO \citep{rafailov2023direct}, this variant computes a preference loss relative to a reference model $\refparams$. The preference loss is the negative log-likelihood of the preference, modeled by the difference between the log-probability ratios:
\begin{equation}
	\loss_{\text{pref}}^{\text{DPO}}(\mathbf{x}_i, p, d) = -\log \sigma \left( \beta \left( \log \frac{\model(\mathbf{x}_i; \params_p)}{\model(\mathbf{x}_i; \refparams_p)} - \log \frac{\model(\mathbf{x}_i; \params_d)}{\model(\mathbf{x}_i; \refparams_d)} \right) \right).
	\label{eq:loss_p_dpo_pref_term}
\end{equation}
The full DPO-based loss for the privileged group, $\loss_{\privileged}^{\text{DPO}}$, combines this preference loss with the BCE fallback:
\begin{align}
	\loss_{\privileged}^{\text{DPO}} = \E_{(\mathbf{x}_i,l) \text{ s.t. } l \in \privileged}
	\left[
	\mathbf{1}_{\confusing \neq \emptyset} \cdot \loss_{\text{pref}}^{\text{DPO}}(\mathbf{x}_i, p, d) + \mathbf{1}_{\confusing = \emptyset} \cdot \baseloss_{\text{BCE}}(\mathbf{x}_i, y_{il}; \params_l)
	\right],
	\label{eq:loss_p_dpo_combined}
\end{align}
where for each sampled privileged label $l$ with a non-empty confusing set, a confusing counterpart $k$ is sampled from $\confusing$, and the pair $(p, d)$ is determined based on $(l, k)$ and the ground truth to compute the loss.

\paragraph{FairPO-SimPO Variant (Reference-Free with Margin).}
This variant adapts SimPO \citep{meng2024simposimplepreferenceoptimization} to create a reference-free preference loss that incorporates an explicit target margin $\gamma > 0$. We adapt SimPO's core concept, which was originally designed for sequences, to our per-label score setting. The resulting preference loss is:
\begin{equation}
	\loss_{\text{pref}}^{\text{SimPO}}(\mathbf{x}_i, p, d) = -\log \sigma \left( \beta \log \frac{\model(\mathbf{x}_i; \params_p)}{\model(\mathbf{x}_i; \params_d)} - \gamma \right).
	\label{eq:loss_p_simpo_pref_term}
\end{equation}
The term $\beta \log(\cdot)$ captures the relative preference between the scores, while the margin term $-\gamma$ enforces a stronger separation, pushing the model to not just rank the labels correctly but to do so by a significant amount. This preference loss is used in place of the DPO term within the overall conditional objective for the \emph{privileged} group (analogous to Eq.~\ref{eq:loss_p_dpo_combined}).

\paragraph{FairPO-CPO Variant (Reference-Free with BCE Regularization).}
Our final and most effective variant adapts CPO \citep{xu2024contrastivepreferenceoptimizationpushing}. This approach is unique in that it integrates a BCE regularizer directly into the preference objective to ensure model stability and a sense of absolute correctness. When a confusing set is found, the preference loss is defined as:
\begin{equation}
	\loss_{\text{pref}}^{\text{CPO}}(\mathbf{x}_i, p, d, l) = -\log \sigma \left( \beta \log \frac{\model(\mathbf{x}_i; \params_p)}{\model(\mathbf{x}_i; \params_d)} \right) + \lambda_{\text{CPO}} \cdot \baseloss_{\text{BCE}}(\mathbf{x}_i, y_{il}; \params_l).
	\label{eq:loss_p_cpo_pref_integrated}
\end{equation}
For each sampled label $l$, the pair $(p, d)$ is determined from a confusing counterpart $k \in \confusing$ if the set is non-empty. Here, the first term is a margin-free preference objective, while the second term, weighted by a hyperparameter $\lambda_{\text{CPO}}$, is the standard BCE loss for the privileged label $l$. This BCE component is crucial, acting as a regularizer that grounds the model in learning absolute scores. The full privileged loss objective is then analogous to the DPO formulation (Eq.~\ref{eq:loss_p_dpo_combined}): the comprehensive $\loss_{\text{pref}}^{\text{CPO}}$ is used when a confusing set is found, and only the standard BCE term (Eq.~\ref{eq:bce_privileged}), is used as a fallback otherwise.

\paragraph{Advantages of the Preference-Based Objective.}
Our choice to adapt modern preference optimization techniques is motivated by the fundamental limitations of standard point-wise losses like \textbf{BCE} and \textbf{Focal Loss} for fine-grained discrimination. These losses evaluate each label's score independently, answering the question, \emph{``Is the score for label $l$ correct in isolation?"}. While effective for general classification, this approach provides an indirect and often insufficient signal for resolving confusion between two closely-scored labels \citep{lin2018focallossdenseobject}. For example, a model might correctly learn to assign a high score (e.g., 0.8) to a true positive label and a slightly lower score (e.g., 0.7) to a confusing negative. A point-wise loss would provide only a small error signal for this negative case. The preference-based objective in FairPO is fundamentally different. It operates on \textit{pairs} of labels and directly answers a more relational question: \emph{``Is the score for the correct label $l$ decisively higher than the score for its specific confusing competitor $k$?"}. By optimizing the log-ratio of these scores, our framework generates a strong, targeted gradient to explicitly drive their values apart. This provides a far more direct and effective mechanism for resolving the most challenging discriminative cases within the \emph{privileged} group, a capability that point-wise losses lack by design.

\subsection{Objective for the Non-Privileged Group: Constrained Performance}
\label{subsec:nonprivileged_objective}

For the \emph{non-privileged} group $\nonprivileged$, the primary objective is different. While the model dedicates its capacity to improving the challenging \emph{privileged} labels, we must ensure that this focus does not come at an unacceptable cost to the performance on the remaining labels. Therefore, our goal for this group is not to aggressively maximize performance, but to \textit{preserve} it by preventing significant degradation relative to a reliable baseline.

To achieve this, we introduce a constrained objective that leverages a pre-trained reference model with parameters $\refparams$. We use the standard BCE loss (Eq.~\ref{eq:bce_privileged}), as our objective for a given label $j \in \nonprivileged$. The model is penalized only if the BCE loss of the current model, $\baseloss_{\text{BCE}}(\mathbf{x}_i, y_{ij}; \params_j)$, exceeds the loss of the reference model, $\baseloss_{\text{BCE}}(\mathbf{x}_i, y_{ij}; \refparams_j)$, by more than a small, predefined slack margin $\epsilon \ge 0$. This is implemented using a hinge loss mechanism:
\begin{equation}
    \loss_{\nonprivileged} = \E_{(\mathbf{x}_i,j) \text{ s.t. } j \in \nonprivileged} \left[ \max(0, \baseloss_{\text{BCE}}(\mathbf{x}_i, y_{ij}; \params_j) - \baseloss_{\text{BCE}}(\mathbf{x}_i, y_{ij}; \refparams_j) - \epsilon) \right].
    \label{eq:loss_np_constrained}
\end{equation}

The gradient for any non-privileged label is zero as long as its performance is \emph{good enough} (i.e., close to or better than the reference model). A learning signal is generated only when performance on a label $j$ drops below this safety threshold.

\subsection{Group Robust Optimization Formulation}
\label{subsec:grpo_formulation}

Having defined two distinct objectives, a preference-based loss $\loss_{\privileged}$ for the \emph{privileged} group and a constrained loss $\loss_{\nonprivileged}$ for the \emph{non-privileged} group, the final challenge is to balance them. As these objectives are often in competition, a simple weighted sum is insufficient. Instead, we formulate the overall training objective as a minimax game, adapting the principles of Group Robust Preference Optimization (GRPO) \citep{ramesh2024group}. We treat the \emph{privileged} and \emph{non-privileged} label sets as two distinct groups. The goal is to train a model, parameterized by $\params$, that is robust to the worst-case distribution of losses across these groups. This is expressed as the following minimax problem:
\begin{equation}
    \min_{\{\params_t\}} \max_{\alpha \in \Delta} \left[ \alpha_{\privileged} \loss_{\privileged} + \alpha_{\nonprivileged} \loss_{\nonprivileged} \right],
    \label{eq:FairPO_objective}
\end{equation}
where $\loss_{\privileged}$ and $\loss_{\nonprivileged}$ represent the expected losses for their respective groups, and $\alpha = (\alpha_{\privileged}, \alpha_{\nonprivileged})$ is a vector of adaptive weights in the probability simplex $\Delta$. This formulation provides a principled way to manage the performance trade-off between the two groups.

\section{Extended Related Works}
\label{app:related_works_extended}

\textbf{Fairness in Multi-Label Classification:}
Ensuring fairness in MLC \citep{zhang2021survey, tantithamthavorn2018impact} is complex due to multi-faceted label relationships. Recent efforts address MLC-specific fairness challenges, such as tackling label imbalance impacting tail labels \citep{wu2023longtailed}, learning instance and class-level subjective fairness \citep{gao2023learning}, or incorporating fairness in dynamic learning settings like class-incremental MLC \citep{masud2023fairnessaware}. FairPO contributes by explicitly partitioning labels into privileged ($\mathcal{P}$) and non-privileged ($\bar{\mathcal{P}}$) sets, applying distinct fairness-motivated objectives to each—notably using preference signals for $\mathcal{P}$—and managing them via a robustness framework. This targeted approach to enhancing performance for pre-defined critical labels, while safeguarding others, differentiates our work.

\textbf{Preference Optimization:}
Preference optimization, especially for aligning LLMs \citep{ouyang2022training, christiano2017deep}, has rapidly advanced. Direct Preference Optimization (DPO) \citep{rafailov2023direct} and its reference-free variants like CPO \citep{xu2024contrastivepreferenceoptimizationpushing} and SimPO \citep{meng2024simposimplepreferenceoptimization} optimize policies directly from preference pairs. The field continues to evolve with methods like Identity Preference Optimization (IPO) for stability \citep{azhar2023ipo}, Kahneman-Tversky-based optimization (KTO) \citep{ethayarajh2024kto}, simple yet effective Rejection Sampling Fine-Tuning (RFT) \citep{dong2023rejection}, and ongoing theoretical analyses of these methods \citep{zhao2024whyisdpo}. We use preferences not to rank entire outputs, but to specifically differentiate true label scores from those of their \textit{confusing} positive/negative counterparts within the privileged label set, thereby sharpening the model's decision boundaries for critical distinctions.

\textbf{Group Robust Optimization:}
Distributionally Robust Optimization (DRO) \citep{ben2009robust}, particularly Group DRO \citep{sagawa2019distributionally}, aims to improve worst-case performance across predefined data groups, enhancing fairness and robustness. This concept sees continued development, for instance, in improving its practicality \citep{zhang2022fromaverage}. Group Robust Preference Optimization (GRPO) \citep{ramesh2024group} extended this to LLM preference learning, balancing performance across preference groups. FairPO directly employs GRPO's adaptive optimization strategy. However, our groups are defined by the label partition ($\mathcal{P}$ and $\bar{\mathcal{P}}$), and GRPO balances their distinct, custom-formulated loss objectives. This provides a principled mechanism for robustly managing the specific fairness-performance trade-offs in our MLC context.

\section{Adapting FairPO for Multi-Attribute Generation}
\label{app:fairpo_gen_future_work}

This section outlines our planned extension of FairPO to multi-attribute generation, a conceptual direction for future work. The goal is to generate a sequence $y$ from a prompt $x$ using a policy $\policy_{\params}(y|x)$ that aligns with fairness goals over a set of attributes $\attributes$. The core idea involves partitioning $\attributes$ into privileged $\privileged$ and non-privileged $\nonprivileged$ sets and retaining the GRPO minimax structure (Eq. \ref{eq:FairPO_objective}). The group losses would be defined over a preference dataset $\dataset_{pref} = \{ (x_i, y_{wi}, y_{li}, j_i) \}_{i=1}^M$, with preference losses like DPO applied to the log-probabilities of entire generated sequences rather than individual label scores.

\par{\textbf{Proposed Privileged Loss} ($\loss_{\privileged}$):}
For privileged attributes $j \in \privileged$, the goal is to ensure the learned policy $\policy_{\params}$ strongly reflects preferences $y_w \succ y_l$ established by that attribute. This is achieved using a standard DPO loss, averaged over the privileged subset of the preference data:
\begin{equation}
	\loss_{\privileged}(\policy_{\params}, \refpolicy) = \E_{(x, y_w, y_l, j) \sim \dataset_{pref} \mid j \in \privileged} \left[ -\log \sigma \left( \beta \cdot h_{\policy_{\params}}(x, y_w, y_l) \right) \right]
	\label{eq:loss_p_gen_future}
\end{equation}
Minimizing this loss directly encourages the model to favor preferred sequences for preferences driven by privileged attributes, relative to the reference policy $\refpolicy$.

\par{\textbf{Proposed Non-Privileged Loss ($\loss_{\nonprivileged}$):}}
For non-privileged attributes $k \in \nonprivileged$, the objective remains analogous to the classification setting: preventing significant performance degradation. This is accomplished with a hinge formulation based on the DPO loss:
\begin{equation}
	\loss_{\nonprivileged}(\policy_{\params}, \refpolicy) = \E_{(x, y_w, y_l, k) \sim \dataset_{pref} \mid k \in \nonprivileged} \left[ \max \left( 0, \loss_{DPO}(\policy_{\params}, \refpolicy; x, y_w, y_l) - (\log 2) - \epsilon' \right) \right].
	\label{eq:loss_pbar_gen_future}
\end{equation}
This penalizes the model only if its preference modeling for non-privileged attributes degrades substantially beyond baseline performance (represented by $\log 2$ for random preference) plus a slack $\epsilon'$. The overall FairPO objective would then use GRPO to balance these two losses.

\section{FairPO Algorithm}
\label{app:FairPO_algo}
The FairPO framework is trained iteratively to solve the minimax objective presented in Eq.~\ref{eq:FairPO_objective}. The detailed procedure, which is inspired by the DPO-based variant of FairPO, is provided in Algorithm~ \ref{alg:grpo_dpo_clf}.

\textbf{Initialization:} The training process begins by initializing the model parameters $\Params$, for instance by copying them from a pre-trained reference model $\Refparams$. The adaptive group weights, $\alpha_{\privileged}$ and $\alpha_{\nonprivileged}$, are typically set to uniform values such as 0.5 each.

\textbf{Iterative Training Loop:} The core of the framework is an iterative training loop. In each step, an instance $(x_i, y_i)$ is sampled from the dataset $\dataset$, and a single label $r \in \mathcal{T}$ is randomly selected from that instance for processing. The subsequent steps depend on whether this sampled label belongs to the privileged or non-privileged set.

If the sampled label $r$ is in the \textbf{privileged set $\privileged$}, the algorithm first identifies if a \textit{confusing set} $S_{il}$ exists for that label (where $l=r$), as detailed in Algorithm~\ref{alg:grpo_dpo_clf}. The loss computation is then conditional on this set:
\begin{itemize}
	\item If confusing examples exist ($S_{il} \neq \emptyset$), a DPO-inspired preference loss is computed between label $l$ and a randomly sampled confusing example $k \in S_{il}$. This preference loss directly encourages the model to improve its ranking of $l$ relative to its specific confounder $k$.
	\item If no confusing examples are found ($S_{il} = \emptyset$), the algorithm reverts to a standard base classification loss (e.g., BCE, Eq.~\ref{eq:bce_privileged}) for label $l$. This fallback is crucial as it ensures the model continues to receive a learning signal on \emph{easier} instances, promoting stable training.
\end{itemize}
The loss calculated from either of these cases contributes to the current step's privileged group loss, $\loss_{\privileged}^{(s)}$. Conversely, if the sampled label $r$ belongs to the \textit{non-privileged set $\nonprivileged$}, the constrained loss $\loss_{\nonprivileged}^{(s)}$ is computed according to Eq.~\ref{eq:loss_np_constrained}. This loss penalizes the model only if its performance on the label $j=r$ deviates from the reference model's performance by more than a predefined slack margin $\epsilon$.

After the appropriate group loss is computed, the GRPO mechanism performs two key updates. First, the \textit{Adaptive Weight Update} adjusts the group weights $\alpha_{\privileged}$ and $\alpha_{\nonprivileged}$ using a mirror ascent step. This step uses an exponential weighting based on the current (and scaled) group losses, dynamically increasing the focus on the group that is currently performing worse (Lines 39-41). Second, the \textit{Model Parameter Update} updates all model parameters $\params_t$ via a mirror descent step, using a combined gradient that is weighted by the newly updated adaptive weights $\alpha_{\privileged}$ and $\alpha_{\nonprivileged}$.

This entire process repeats for a predefined number of iterations or until convergence, allowing FairPO to dynamically balance its objectives to achieve robust fairness. For variants like FairPO-SimPO or FairPO-CPO, the core logic remains identical; only the DPO-inspired preference loss component is replaced with their respective preference formulations (e.g., Eq.~\ref{eq:loss_p_simpo_pref_term} or \ref{eq:loss_p_cpo_pref_integrated}). The overall GRPO structure and non-privileged handling are consistent across all variants.

\begin{algorithm}[p]
	\caption{FairPO Algorithm for Multi-Label Classification (DPO-inspired)}
	\label{alg:grpo_dpo_clf}
	\begin{algorithmic}[1]
		\State \textbf{Initialize:} $\{\mathbf{w}_t^{(0)} \in \R^d \vert \forall t \in \mathcal{T}\}$ (e.g., copy $\{\hat{\mathbf{w}}_t \vert \forall t \in \mathcal{T}\}$), $\alpha_{\privileged}^{(0)} \leftarrow 0.5, \alpha_{\nonprivileged}^{(0)} \leftarrow 0.5$.
		\State \textbf{Choose:} $\eta_{\params}, \eta_\alpha$, $\beta$, $\{\hat{\mathbf{w}}_t \vert \forall t \in \mathcal{T}\}$, $\epsilon$.
		\State \textbf{for} $s = 0$ to $S$ (MaxIterations) \textbf{do}
		\State \hspace{0.5cm} Sample an example: $(x_i, [y_{i1}, \dots, y_{iT}]) \in \dataset \sim p_{\dataset}(.)$.
		\State \hspace{0.5cm} Initialize group losses for this step: $\loss_{\privileged}^{(s)} \leftarrow 0$, $\loss_{\nonprivileged}^{(s)} \leftarrow 0$.
		\State \hspace{0.5cm} Initialize gradients: $g^t_{\privileged} \leftarrow \vec{0}$, $g^t_{\nonprivileged} \leftarrow \vec{0} \quad \forall t \in \mathcal{T}$.
		\State \hspace{0.5cm} Forward pass: $m(x_i;\params_t^{(s)}) \leftarrow \sigma(\params_t^{(s)^T}\mathbf{z}_i)$ where $\mathbf{z}_i \leftarrow \pi_{\theta}(x_i) \quad \forall t \in \mathcal{T}$.
		\State \hspace{0.5cm} Sample a label: $r \in \mathcal{T}\sim Uniform(\frac{1}{|\mathcal{T}|})$.
		
		\State \hspace{0.5cm} \textbf{if} $r \in \privileged$ \textbf{then} \Comment{Handle privileged label}
		\State \hspace{1.0cm} $l \leftarrow r$, $S_{il}^{\text{neg}} \leftarrow \emptyset$, $S_{il}^{\text{pos}} \leftarrow \emptyset$
		\State \hspace{1.0cm} \textbf{if} $y_{il} = +1$ \textbf{then} \Comment{True Positive case for privileged label $l$}
		\State \hspace{1.5cm} $S_{il}^{\text{neg}} \leftarrow \{ k \in \labels \mid y_{ik} = 0 \text{ and } \model(x_i; \params_k^{(s)}) \ge \model(x_i; \params_l^{(s)}) \}$, $\confusing \leftarrow S_{il}^{\text{neg}}$
		\State \hspace{1.0cm} \textbf{else if} $y_{il} = 0$ \textbf{then} \Comment{True Negative case for privileged label $l$}
		\State \hspace{1.5cm} $S_{il}^{\text{pos}} \leftarrow \{ k \in \labels \mid y_{ik} = +1 \text{ and } \model(x_i; \params_k^{(s)}) \le \model(x_i; \params_l^{(s)}) \}$, $\confusing \leftarrow S_{il}^{\text{pos}}$
		\State \hspace{1.0cm} \textbf{end if}
		
		\State \hspace{1.0cm} \textbf{if} $\confusing \neq \emptyset$ \textbf{then} \Comment{Confusing examples exist, use DPO-inspired loss}
		\State \hspace{1.5cm} Sample $k \in \confusing \sim Uniform(\frac{1}{|\confusing|})$
		\State \hspace{1.5cm} \textbf{if} $y_{il} = +1$ \textbf{then} \Comment{DPO for True Positive $l$ vs Confusing Negative $k$}
		\State \hspace{2.0cm} $h_{\params^{(s)}}(x_i, l, k) \leftarrow \left( \log  \frac{\model(x_i; \params_l^{(s)})}{\model(x_i; \refparams_l)} \right) - \left( \log  \frac{\model(x_i; \params_k^{(s)})}{\model(x_i; \refparams_k)} \right)$.
		\State \hspace{2.0cm} $\loss_{\text{pref}} \leftarrow  -\log \sigma \left( \beta \cdot h_{\params^{(s)}}(x_i, l, k) \right)$
		\State \hspace{1.5cm} \textbf{else if} $y_{il} = 0$ \textbf{then} \Comment{DPO for True Negative $l$ vs Confusing Positive $k$}
		\State \hspace{2.0cm} $h_{\params^{(s)}}(x_i, k, l) \leftarrow \left( \log  \frac{\model(x_i; \params_k^{(s)})}{\model(x_i; \refparams_k)} \right) - \left( \log  \frac{\model(x_i; \params_l^{(s)})}{\model(x_i; \refparams_l)} \right)$.
		\State \hspace{2.0cm} $\loss_{\text{pref}} \leftarrow  -\log \sigma \left( \beta \cdot h_{\params^{(s)}}(x_i, k, l) \right)$
		\State \hspace{1.5cm} \textbf{end if}
		\State \hspace{1.5cm} $\loss_{\privileged}^{(s)} \leftarrow \loss_{\text{pref}}$, $g_{\privileged}^t \leftarrow g_{\privileged}^t + \nabla_{\params_t} \loss_{\text{pref}} |_{\params_t^{(s)}} \quad \forall t \in \mathcal{T}$.
		\State \hspace{1.0cm} \textbf{else} \Comment{No confusing examples, use BCE loss for privileged label $l$}
		\State \hspace{1.5cm} $\loss_{\text{BCE}} \leftarrow -[y_{il} \log \model(x_i; \params_l^{(s)}) + (1-y_{il}) \log(1-\model(x_i; \params_l^{(s)}))]$
		\State \hspace{1.5cm} $\loss_{\privileged}^{(s)} \leftarrow \loss_{\text{BCE}}$, $g_{\privileged}^t \leftarrow g_{\privileged}^t + \nabla_{\params_t} \loss_{\text{BCE}} |_{\params_t^{(s)}} \quad \forall t \in \mathcal{T}$.
		\State \hspace{1.0cm} \textbf{end if}
		
		\State \hspace{0.5cm} \textbf{else if} $r \in \nonprivileged$ \textbf{then} \Comment{Handle non-privileged label}
		\State \hspace{1.0cm} $j \leftarrow r$
		\State \hspace{1.0cm} $\baseloss(\params_j^{(s)}) \leftarrow -[y_{ij} \log (\model(x_i;\params_j^{(s)})) + (1-y_{ij})\log (1-\model(x_i;\params_j^{(s)}))]$
		\State \hspace{1.0cm} $\baseloss(\refparams_j) \leftarrow -[y_{ij} \log (\model(x_i;\refparams_j)) + (1-y_{ij})\log (1-\model(x_i;\refparams_j))]$
		\State \hspace{1.0cm} $\loss_{\nonprivileged}^{(s)} \leftarrow \max \left( 0, \, \baseloss(\params_j^{(s)}) - \baseloss(\refparams_j) - \epsilon \right)$, $g_{\nonprivileged}^t \leftarrow g_{\nonprivileged}^t + \nabla_{\params_t} \loss_{\nonprivileged}^{(s)} |_{\params_t^{(s)}} \quad \forall t \in \mathcal{T}$.
		\State \hspace{0.5cm} \textbf{end if}
		
		\State \hspace{0.5cm} $\alpha_{\privileged}^{(s+1)} \leftarrow \alpha_{\privileged}^{(s)} \exp(\eta_\alpha \loss_{\privileged, scaled}^{(s)})$ and $\alpha_{\nonprivileged}^{(s+1)} \leftarrow \alpha_{\nonprivileged}^{(s)} \exp(\eta_\alpha \loss_{\nonprivileged, scaled}^{(s)})$ \Comment{Mirror ascent} 
		\State \hspace{0.5cm} $Z \leftarrow \alpha_{\privileged}^{(s+1)} + \alpha_{\nonprivileged}^{(s+1)}$, $\alpha_{\privileged}^{(s+1)} \leftarrow \frac{\alpha_{\privileged}^{(s+1)}}{Z}$ and $\alpha_{\nonprivileged}^{(s+1)} \leftarrow \frac{\alpha_{\nonprivileged}^{(s+1)}}{Z}$ \Comment{Weight normalization}
		\State \hspace{0.5cm} $\params_t^{(s+1)} \leftarrow \params_t^{(s)} - \eta_{\params} (\alpha_{\privileged}^{(s+1)} g_{\privileged}^t + \alpha_{\nonprivileged}^{(s+1)} g_{\nonprivileged}^t)$ \quad $\forall t \in \mathcal{T}$\Comment{Mirror descent}
		\State \textbf{end for}
		\State \textbf{return} $\{\mathbf{w}_t^{(S)} \vert \forall t \in \mathcal{T}\}$
	\end{algorithmic}
\end{algorithm}

\section{Dataset and Preprocessing Details}
\label{app:dataset_details}

\textbf{MS-COCO 2014} \citep{lin2015microsoftcococommonobjects}: We used the official 2014 train/val splits. The training set contains 82,783 images and the validation set (used as our test set) contains 40,504 images. There are 80 object categories. The privileged set $\mathcal{P}$ consisted of the 16 labels (20\% of 80) with the lowest frequency in the training set. The remaining 64 labels formed $\nonprivileged$.

\textbf{NUS-WIDE} \citep{chua2009nus}: This dataset contains 269,648 images with 81 concept labels. We used the common split of 161,789 images for training and 107,859 for testing. The privileged set $\mathcal{P}$ consisted of the 16 labels (approx. 20\% of 81) with the lowest frequency in the training set. The remaining 65 labels formed $\nonprivileged$.

\textbf{Image Preprocessing}: For both datasets, images were resized to $224 \times 224$ pixels and normalized using the standard ImageNet mean and standard deviation, consistent with the ViT pretraining. Standard data augmentations like random horizontal flips and random resized crops were applied during training.

\section{FairPO Experimental Details}
\label{app:FairPO_exp_details}

\subsection{Common Setup for All FairPO Variants}
Unless specified otherwise, a common setup was used for all FairPO variants to ensure fair comparison. The base model for feature extraction was a Vision Transformer (ViT), specifically \texttt{vit-base-patch16-224} \citep{dosovitskiy2021imageworth16x16words}, which was pre-trained on ImageNet-21k and fine-tuned on ImageNet-1k. During our fine-tuning, the ViT backbone was kept frozen, with the exception of its final encoder layer, which was made trainable to allow for adaptation of higher-level features. All experiments were conducted on the MS-COCO 2014 \citep{lin2015microsoftcococommonobjects} and NUS-WIDE \citep{chua2009nus} datasets. Images were resized to $224 \times 224$ pixels, normalized using ImageNet statistics, and augmented with standard techniques like random horizontal flips and resized crops. The AdamW optimizer \citep{loshchilov2019decoupled} was used to update all trainable parameters. Models were trained for a maximum of 25 epochs with a batch size of 32, and we employed an early stopping strategy with a patience of 5 epochs based on the overall mAP on the validation set.

\subsection{Per-Label Non-Linear MLP Classifier Head}
\label{app:mlp_head_details}
For each of the $T$ labels in a dataset, we employed a dedicated and independent non-linear Multi-Layer Perceptron (MLP) head to predict the probability of that label being positive. Using separate MLP heads allows for more complex, non-linear decision boundaries tailored to each label's specific characteristics, which is particularly beneficial for labels with varying difficulty. Each MLP head takes the $d$-dimensional feature vector (where $d=768$ for ViT-Base) from the ViT's [CLS] token as input and outputs a single logit. The final probability score $\model(x_i; \params_t)$ is obtained by applying a sigmoid function to this logit. The specific architecture for each MLP head is as follows:
\begin{enumerate}[noitemsep]
	\item Linear Layer: $d \rightarrow 256$ neurons, followed by ReLU Activation
	\item Linear Layer: $256 \rightarrow 64$ neurons, followed by ReLU Activation
	\item Linear Layer: $64 \rightarrow 16$ neurons, followed by ReLU Activation
	\item Linear Layer: $16 \rightarrow 4$ neurons, followed by ReLU Activation
	\item Linear Layer (Output): $4 \rightarrow 1$ neuron (producing the logit)
\end{enumerate}
The parameters $\params_t$ for each label $t$'s MLP head are unique to that label. All parameters within these MLP heads were fully trainable during both the SFT pre-training (for the reference model) and the final FairPO fine-tuning.

\end{document}